\documentclass[11pt]{article}

\usepackage{deauthor}
\usepackage{times,graphicx}
\usepackage[square,comma,numbers,sort]{natbib}
\usepackage{url}
\usepackage{hyperref}
\usepackage{multirow,multicol,xspace}
\usepackage{amsmath,amssymb}
\usepackage{float}
\usepackage{booktabs}
\usepackage{graphics}
\usepackage{subfigure}
\usepackage{forest}
\usepackage[capitalize,noabbrev]{cleveref}

\definecolor{mydarkblue}{rgb}{0,0.08,0.45}
\definecolor{myblue}{HTML}{3b75c3}
\definecolor{myred}{HTML}{E33222}
\definecolor{mygreen}{HTML}{438773}
\definecolor{mymaroon}{RGB}{142,27,19}
\definecolor{maroon}{HTML}{800000}
\definecolor{mycite}{cmyk}{0.55,1,0,0.15}
\definecolor{codeblue}{rgb}{0.25,0.5,0.5}
\definecolor{codekw}{rgb}{0.85, 0.18, 0.50}
\definecolor{codegreen}{rgb}{0,0.6,0}
\definecolor{codegray}{rgb}{0.5,0.5,0.5}
\definecolor{codepurple}{rgb}{0.58,0,0.82}
\definecolor{backcolour}{rgb}{0.95,0.95,0.92}
\hypersetup{
    colorlinks=true,
    citecolor=codeblue,
    linkcolor=mymaroon,
    urlcolor=maroon,
    bookmarks=true,
    bookmarksnumbered=true,
    pdfpagemode={UseOutlines},
    plainpages=false,
    pdfpagelabels=true,
    pdftitle={Graph Data Augmentation for Graph Machine Learning: A Survey},
    pdfsubject={Graph Data Augmentation for Graph Machine Learning: A Survey},
    pdfauthor={T. Zhao, et al.}
}

\usepackage{pifont}
\newcommand{\cmark}{\ding{51}}%

\renewcommand\paragraph[1]{\vspace{0.05in} \noindent \textbf{#1.}}
\newcommand*{\Scale}[2][4]{\scalebox{#1}{$#2$}}

\bulletinyear{2023}

\begin{document}
\title{Graph Data Augmentation for Graph Machine Learning:\\ A Survey}
\author
{Tong Zhao$^{1,4}$, Wei Jin$^2$, Yozen Liu$^1$, Yingheng Wang$^3$, Gang Liu$^4$, \\ 
Stephan Günnemann$^5$, Neil Shah$^1$, Meng Jiang$^4$ \\
\small{$^1$Snap Inc., $^2$Michigan State University, $^3$Cornell University,} \\
\small{$^4$University of Notre Dame, $^5$Technical University of Munich} \\
\small\texttt{$^1$\{tzhao,yliu2,nshah\}@snap.com,$^2$jinwei2@msu.edu,$^3$yw2349@cornell.edu,}\\
\small\texttt{$^4$\{gliu7,mjiang2\}@nd.edu,$^5$guennemann@in.tum.de}
}

\maketitle

\begin{abstract}
Data augmentation has recently seen increased interest in graph machine learning given its demonstrated ability to improve model performance and generalization by added training data. Despite this recent surge, the area is still relatively under-explored, due to the challenges brought by complex, non-Euclidean structure of graph data, which limits the direct analogizing of traditional augmentation operations on other types of image, video or text data.  Our work aims to give a necessary and timely overview of existing graph data augmentation methods; notably, we present a comprehensive and systematic survey of graph data augmentation approaches, summarizing the literature in a structured manner. We first introduce three different taxonomies for categorizing graph data augmentation methods from the data, task, and learning perspectives, respectively. Next, we introduce recent advances in graph data augmentation, differentiated by their methodologies and applications. We conclude by outlining currently unsolved challenges and directions for future research. Overall, our work aims to clarify the landscape of existing literature in graph data augmentation and motivates additional work in this area, providing a helpful resource for researchers and practitioners in the broader graph machine learning domain.  Additionally, we provide a continuously updated reading list at \url{https://github.com/zhao-tong/graph-data-augmentation-papers}.

\end{abstract}

\section{Introduction}
\label{sec:intro}
\noindent Data driven inference has received a significant boost in generalization capability and performance improvement in recent years from data augmentation (DA) techniques. DA techniques increase the amount of training data by creating plausible variations of existing data without additional ground-truth labeling efforts, and have seen widespread adoption in fields such as computer vision (CV)~\cite{cubuk2019autoaugment} and natural language processing (NLP)~\cite{feng2021survey}. These techniques allow machine learning models to learn to generalize across those variations and attend to signal over noise.

In recent years, with the rapid development of graph machine learning (GML) methods such as graph neural networks (GNNs)~\cite{kipf2016semi,hamilton2017inductive}, studies have shown that the effectiveness of GML approaches also largely depends on the data quality. Given the dependent nature of graph data and the message-passing design of most GNNs, GML faces unique challenges such as: structural data sparsity brought by power-law degree distributions in most graphs, noisy and even erroneous topology brought by imperfect construction of the graph structure from raw data under other formats, low quality and incomplete node attributes, adversarial attacks on structure and attributes, lack of labelled data due to costly human annotations, and over-smoothing caused by the message passing design in GNNs.
As DA allows researchers to alleviate such challenges from a data perspective, there has been increased interest and demand for such techniques on graph data~\cite{zhao2021data}, and there has been a growing number of works on graph data augmentation (GDA).

With the irregular and non-Euclidean structure of graph data, it is non-trivial to directly analogize DA techniques from CV and NLP to the graph domain, except for the most basic operations such as random masking/dropping/cropping. To better promote the effectiveness of GML approaches and alleviate the unique challenges in GML, recent literature designed graph-specific augmentation techniques following methodologies such as graph structure learning, graph adversarial training, graph rationalization, etc.
Creating a unified taxonomy for all GDA techniques is not intuitive as they can be categorized under different facets. 
For example, taking the data modelity that the augmentation methods work on, they can be separated into structure augmentations, feature augmentations, and label augmentations. On the other hand, the focusing downstream tasks (i.e., node-level, edge-level, and graph-level tasks) can also categorize the GDA techniques. Moreover, the GDA methods can also be separated by whether the methods involves learning during the augmentation process. That is, whether they are rule-based approaches or learned approaches.

This paper aims to sensitize the GML community towards this growing area of work, as DA has already drawn much attention in CV and NLP~\cite{cubuk2019autoaugment,feng2021survey}.
As interest and work on this topic continue to increase, this is an opportune time for a comprehensive work to (i) introduce background and motivation of GDA, (ii) give a bird's-eye view of existing GDA techniques under different taxonomies, (iii) introduce representative GDA techniques with their usage and applications, and (iv) identify key challenges to effectively motivate and orient interest in this area.
We hope this survey can serve as a guide for researchers and practitioners who are new to or interested in studying this topic, and also inspire future research in this area.

The text is structured as follows: \cref{sec:background} gives background and motivation on GNNs and GDA. It defines GDA and motivates its use in GML tasks.
\cref{sec:taxonomy} categorizes GDA techniques based on three different taxonomies: the operated graph data, the downstream tasks, and whether the method involves learning. \cref{sec:simpleaug} describes rule-based GDA techniques for GML -- which we partition into Data Removal (\cref{sec:dataremove}), Data Addition (\cref{sec:dataadd}), and Data Manipulation (\cref{sec:datamani}) focuses. Similarly, \cref{sec:learnedaug} introduces learned GDA techniques, which are further categorized by their methodologies: Graph Structure Learning (\cref{sec:gsl}), Graph Adversarial Training (\cref{sec:adv}), Graph Rationalization (\cref{sec:rat}), and Automated Augmentation (\cref{sec:autoaug}).
\cref{sec:application} introduces GDA techniques that are used under three different self-supervised learning objectives: Contrastive Learning (\cref{sec:contrastive}), Non-contrastive Learning (\cref{sec:noncontrastive}), and Consistency Training (\cref{sec:consistent}). Finally, \cref{sec:future} discusses challenges and future directions for GDA.


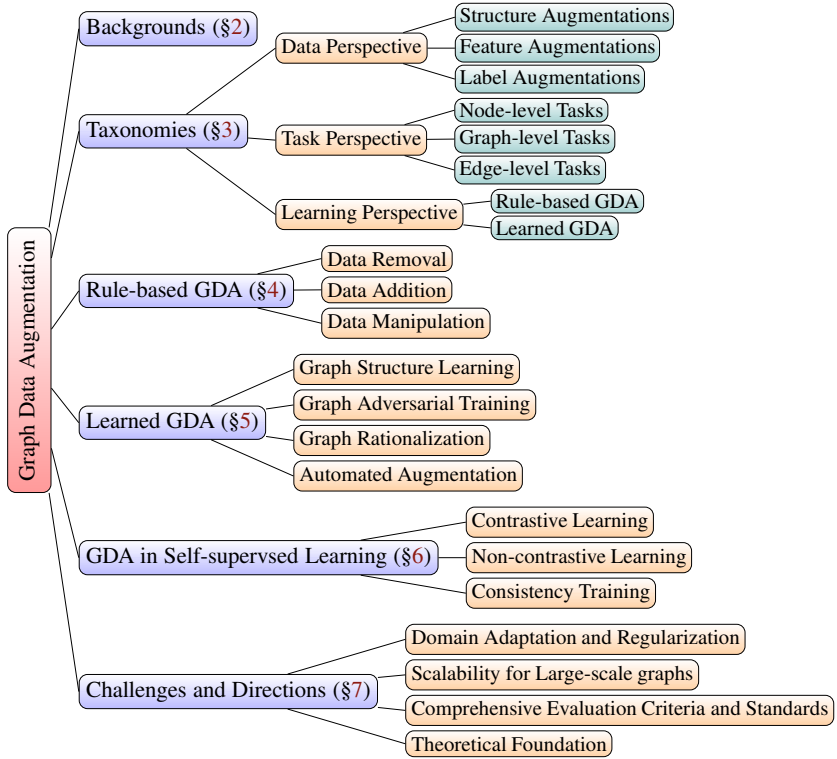
\begin{figure}[!th]
\centering
\resizebox{0.65\linewidth}{!}{
\begin{forest}
	for tree={
		draw,
		shape=rectangle,
		rounded corners,
		top color=white,
		grow'=0,
		l sep'=1.2em,
		reversed=true,
		anchor=west,
		child anchor=west,
	},
	root/.style={
		rotate=90, shading angle=90, bottom color=red!40,
		anchor=north, font=\normalsize, inner sep=0.5em},
	level1/.style={
		bottom color=blue!30, font=\normalsize, inner sep=0.3em,
		s sep=0.2em},
	level2/.style={
		bottom color=orange!40, font=\small, inner sep=0.25em,
		s sep=0.1em},
	level3/.style={
		bottom color=teal!40, font=\small, inner sep=0.2em,
		l sep'=0.5em},
	where n=0{root}{},
	where level=1{level1}{},
	where level=2{level2}{},
	where level=3{level3}{},
	[Graph Data Augmentation
            [Backgrounds (\S \ref{sec:background})]
		[Taxonomies (\S \ref{sec:taxonomy})
                [Data Perspective
                    [Structure Augmentations]
                    [Feature Augmentations]
                    [Label Augmentations]
                ]
                [Task Perspective
                    [Node-level Tasks]
                    [Graph-level Tasks]
                    [Edge-level Tasks]
                ]
                [Learning Perspective
                    [Rule-based GDA]
                    [Learned GDA]
                ]
		]
		[Rule-based GDA (\S \ref{sec:simpleaug})
			[Data Removal]
			[Data Addition]
			[Data Manipulation]
		]
		[Learned GDA (\S \ref{sec:learnedaug})
			[Graph Structure Learning]
			[Graph Adversarial Training]
                [Graph Rationalization]
                [Automated Augmentation]
		]
		[GDA in Self-supervsed Learning (\S \ref{sec:application})
			[Contrastive Learning]
			[Non-contrastive Learning]
			[Consistency Training]
		]
            [Challenges and Directions (\S \ref{sec:future})
                [Domain Adaptation and Regularization]
                [Scalability for Large-scale graphs]
                [Comprehensive Evaluation Criteria and Standards]
                [Theoretical Foundation]
            ]
	]
\end{forest}
}
\caption{Structure of this survey.}
\label{fig:overview}
\end{figure}

\section{Preliminaries}
\label{sec:background}

\subsection{Notations}
Let $G = (\mathcal{V}, \mathcal{E})$ be a graph of $N$ nodes, where $\mathcal{V} = \{v_1, v_2, \dots, v_N\}$ is the set of $N$ nodes and $\mathcal{E} \subseteq \mathcal{V} \times \mathcal{V}$ is the set of links. We denote the adjacency matrix as $\mathbf{A} \in \{0, 1\}^{N \times N}$, where $A_{i,j} = 1$ indicates nodes $v_i$ and $v_j$ are connected and vice versa. We denote the node feature matrix as $\mathbf{X} \in \mathbb{R}^{N \times F}$, where $F$ is the number of raw node features and $\boldsymbol{x}_{i}$ indicates the feature vector of node $v_i$ (the $i$-th row of $\mathbf{X}$). We use $\boldsymbol{y}$ to denote the label each sample, which can be node, edge, or graph depending on the task. We use symbol with tilde to denote the data generated by GDA methods. For example, $\tilde{\mathbf{A}}$ for the augmented adjacency matrix, $\tilde{\boldsymbol{x}}_i$ for the augmented feature vector of node $v_i$,  etc.

\subsection{Graph Neural Networks}

Graph neural networks (GNNs) enjoy widespread use in modern graph-based machine learning due to their flexibility to incorporate node features, custom aggregations, and inductive operation, unlike earlier works which were based on embedding lookups~\cite{perozzi2014deepwalk,grover2016node2vec}. Following the initial idea of convolution based on spectral graph theory~\cite{bruna2013spectral}, many spectral GNNs have since been developed and improved by~\cite{defferrard2016convolutional,kipf2016semi,levie2018cayleynets,klicpera2018predict,ma2021unified}. As spectral GNNs generally operate (expensively) on the full adjacency, spatial-based methods which perform graph convolution with neighborhood aggregation became prominent~\cite{hamilton2017inductive,velivckovic2017graph}, owing to their scalability and flexibility~\cite{ying2018graph,wu2020comprehensive}. 

Generally, the generic formulation of message passing-based GNNs can be defined by an aggregation function (\textsc{Aggregate}) and an update function (\textsc{Update}). In each layer, \textsc{Aggregate} aggregates the embeddings from previous layer for each node from all its neighbors, and \textsc{Update} updates each node's embedding by combining its own previous embedding and the aggregated neighbor embeddings \cite{hamilton2017inductive}. Specifically,
\begin{equation}
\label{eq:gnn}
\begin{split}
    \boldsymbol{h}_{\mathcal{N}(v)}^l &= \textsc{Aggregate}(\{\boldsymbol{h}_u^{l-1} | u \in \mathcal{N}(v) \}), \\
    \boldsymbol{h}_v^l &= \textsc{Update} (\boldsymbol{h}_v^{l-1}, \boldsymbol{h}_{\mathcal{N}(v)}^l),
\end{split}
\end{equation}
where $\boldsymbol{h}_v^l$ denotes the representation vector of node $v$ at the $l$-th GNN layer, and $\mathcal{N}(v)$ denotes the set of neighbors of node $v$.

In implementation, GNNs can usually be implemented with (sparse) matrix multiplications. Without the loss of generality, here we take the most commonly used Graph Convolutional Network (GCN)~\cite{kipf2016semi} as an example. One layer of GCN is defined as
\begin{equation}
\label{eq:gcn}
    \mathbf{H}^l = \sigma(\mathbf{D}^{-\frac{1}{2}}\mathbf{A}\mathbf{D}^{-\frac{1}{2}}\mathbf{W}^l\mathbf{H}^{l-1}),
\end{equation}
where $\mathbf{D}$ is the diagonal degree matrix s.t. $\mathbf{D}_{i,i} = \sum_j \mathbf{A}_{i,j}$ (assuming $\mathbf{A}$ contains self-loops), $\sigma(\cdot)$ is the nonlinear activation function such as ReLU, and $\mathbf{W}^l$ denotes the learnable weight matrix at the $l$-th GNN layer. Furthermore, we use $g_\Theta(\cdot)$ to denote the mapping function of the whole GNN model parameterized by $\Theta$.

\subsection{Graph Data Augmentation}

The DA area encompasses techniques of increasing/generating training data without directly collecting or labeling more data. Most DA techniques either add slightly modified copies of existing data, or generate synthetic data based on existing data. The augmented data act as a regularizer and reduce overfitting when training data-driven models~\cite{shorten2019survey}. DA techniques has been commonly used in CV~\cite{cubuk2019autoaugment} and NLP~\cite{feng2021survey}, where augmentation operations such as cropping, flipping, and back-translation are usually used in machine learning model training. 

In GML, in contrast to regular and Euclidean data such as grids (e.g., images) and sequences (e.g., sentences), the graph structure is encoded by node connectivity, which is non-Euclidean and irregular. Most structured augmentation operations used frequently in CV and NLP cannot be easily analogized to graph data. Therefore, how to design effective augmentations of graph data is less obvious. For example, the data objects for node-level and edge-level tasks are inter-connected and non-i.i.d, meaning that GDA techniques typically modify the entire dataset (graph) instead of a specific data object (nodes or edge) in isolation. Generally, a GDA method can be defined as a transformation function $f: G \rightarrow \tilde{G}$, where the the transformation function $f$ can be either rule-based or learnable, and the augmented graph $\tilde{G}$ contains the augmented adjacency matrix $\tilde{\mathbf{A}}$ and node feature matrix $\tilde{\mathbf{X}}$ (and optionally augmented edge features, node or graph labels). Moreover, the augmentation function $f$ is not necessarily deterministic. That is, the same $f$ may generate multiple different versions of the augmented graph $\tilde{G}$, and the model may use one or multiple of these augmentations as required for training.

\subsection{Motivation: Why Augment Graphs?}
Graphs are often utilized to model or represent an underlying process of relationships or affinities; for example, ``which individuals are friends with one another?'' or ``which movies do individuals like?''  In some cases, these relationships are strictly defined and known, e.g. researchers jointly co-authoring articles, or atoms interacting in a chemical compound.  However, in many other cases, an ``observed'' graph may be misaligned with the true process it intends to model for a variety of reasons \cite{brugere2018network}.  In some cases, like in social interaction graphs, noise may be inadvertently or adversarially introduced by spammers who pollute underlying data about authentic interactions with inauthentic ones for nefarious purposes \cite{shah2014spotting, kumar2018false}.  In other cases, noise may be inherently created by limited or partial observation (e.g. a movie recommendation system never recommending a certain genre of movies to a group of users) caused by privacy reasons \cite{chierichetti2015efficient, duong2011modeling}, biased recommendation policies \cite{joachims2016counterfactual, zhao2021counterfactual}, or other reasons.  Noise can also occur by measurement or thresholding errors (e.g. discretizing continuous signals between brain voxels into discrete ones) \cite{garrison2015stability}, or human errors (e.g. a person forgetting to add a known contact to their phone's contact-book).  All of these scenarios can introduce gaps between an intended and observed graph.  Moreover, even if all relationships a graph intends to capture are observed properly, there is no guarantee that the graph is a particularly useful \cite{brugere2018network} for a particular downstream learning task, especially when utilized in a GML context, e.g. a graph connecting individuals by similar heights may be unhelpful in regressing income.

GDA methods offer an attractive solution in denoising, imputing, and generally enhancing graph structure to align better with an intended modeling processes, or objectives of a target learning task \cite{zhao2021data}.  Adding or removing nodes and edges can help connect or disconnect a graph to facilitate its use towards targeted objectives.  Moreover, utilizing heuristic graph modification strategies to increase model exposure in training may lead to better generalizing, more robust and higher performance models \cite{wang2020nodeaug, kong2020flag, zhou2020data}.  Both learned and rule-based DA techniques have shown immense potential in other domains like tabular ML (e.g. oversampling \cite{barandela2004imbalanced} and SMOTE \cite{chawla2002smote}),  CV (e.g. rotations, flips and translations of images \cite{shorten2019survey} and random erasure \cite{zhong2020random}) and NLP (e.g. synonym replacement and random token additions/deletions \cite{wei2019eda} and back-translation \cite{sennrich2015improving}); however, as aforementioned, these techniques usually lack clear analogs in the graph domain due to unclear correspondence of their label-preserving transforms.  This lack of clarity motivates work into understanding the limitations of graphs, suitable designs for augmentation techniques, and their breadth of impact. 


\section{Taxonomies}
\label{sec:taxonomy}

In this section, we introduce three different taxonomies that can be used to categorize GDA techniques. They come from different perspectives of data, task, and learnability, respectively. As these taxonomies are orthogonal to each other, and each of them can in some way categorize all GDA methods, we will only focus on one taxonomy (rule-based vs. learned augmentation) for the later sections.

\subsection{Operated Data Modality}
As GDA methods all operate on graph data, they can naturally be categorized by the data modality that they aim to manipulate. Therefore, one intuitive taxonomy for GDA methods would be classifying them into one or more of three categories: structure, feature, and label augmentations.

\textbf{Structure Augmentations} 
are the GDA operations that modify the graph connectivity via adding/removing edges or adding/removing nodes from the graph. The modifications can be either deterministic (e.g., GDC~\cite{klicpera2019diffusion} and GAug-M~\cite{zhao2021data} both modify the graph structure and used the modifed graph for training/inferencing) or stochastic (e.g., DropEdge\cite{rong2019dropedge} and DropNode~\cite{feng2020graph} randomly drop edges/nodes from the observed training graph). 
\textbf{Feature Augmentations} 
are the GDA operations that modify or create raw node features. For example, \citet{you2020graph} used Attribute Masking that randomly masked off node features; FLAG~\cite{kong2020flag} augments node features with gradient-based adversarial perturbations. It's worth noting that stucture augmentations and feature augmentations are also sometimes combined in some GDA methods. For example, MoCL~\cite{sun2021mocl} substitutes subgraphs in molecular graphs with subgraphs of different functional groups.
\textbf{Label Augmentations} 
are the GDA operations that involves modifying the labels. For example, Mixup-based methods~\cite{han2022G,guo2021intrusion} interpolate existing training examples and assign new label for the generated example. Counterfactual data augmentation methods (e.g., CFLP~\cite{zhao2021counterfactual}) generate counterfactual examples with corresponding new labels.

\subsection{Downstream Tasks}
Another straightforward taxonomy of categorizing GDA methods is by the downstream tasks that they tackle. Generally, most GML methods can be categorized into three high-level task types: \textbf{node-level}, \textbf{edge-level}, and \textbf{graph-level} tasks. 
Similarly, many GDA methods are designed toward one of these tasks, and cannot be easily generalized to other tasks. For example, CFLP~\cite{zhao2021counterfactual} generates counterfactual links as augmented data specifically for training a neural link predictor, and these counterfactual links are useless to other tasks like node classification as they are counterfactual labels on node pairs under specific treatments. Moreover, certain GDA methods that are designed for molecular graphs (e.g., MoCL~\cite{sun2021mocl}) are not opeartable on the large graph datasets used in other tasks as they rely on the domain specific substructures of molecular graphs, e.g., functional groups. Nonetheless, the downside of categorizing by downstream tasks is that a fair number of GDA methods were designed more generically for various tasks; for example, DropEdge~\cite{rong2019dropedge} simply conducts random edge dropping during training, and the method can naturally be applied on most GML methods.

\subsection{Rule-based vs. Learned Augmentations}

GDA methods can also be categorized by whether the augmentation process involved learning, namely rule-based GDA approaches and learned GDA approaches. More specifically, \textbf{rule-based GDA approaches} refer to the non-learnable methods that modify or manipulate the graph data following pre-defined rules, which can be stochastic, deterministic, or mixture of both. A rule-based GDA method can be as simple as randomly removing a given fraction of edges~\cite{rong2019dropedge} or randomly cropping out part of the graph~\cite{you2020graph}; it can also be more complicated such as counterfactual augmentation~\cite{zhao2021counterfactual} based on similarity matching rules and graph diffusion methods~\cite{klicpera2019diffusion} that follows specific diffusion kernels. We also categorize Mixup-based augmentations~\cite{han2022G} as rule-based approaches since they usually only contain one non-learnable parameter (sampled from pre-defined distributions) when generating new data objects by interpolating two existing data objects. 

On the other hand, \textbf{learned GDA approaches} refer to the augmentation methods that contains learnable parameters in the process of generating augmented examples. The augmentation module can either be trained independently or in an end-to-end style with the downstream classifier or regressor~\cite{zhao2021data}. For example, graph structure learning methods~\cite{zhu2021survey,jin2020graph,zhao2021data} often assume the observed graph data is noisy, incomplete, or entirely missing, so they first try to learn the ``clean'' graph structure before using it in the training and inference of GNNs. Graph rationalization methods~\cite{wu2021discovering,liu2022graph} learn subgraphs that are likely to be causally related with the graph labels and use them for augmentation. Automated augmentation methods~\cite{zhao2022autogda,luo2022automated} utilize reinforcement learning agents to learn the optimal augmentation strategy for the given data automatically. 

In \cref{sec:simpleaug,sec:learnedaug}, we will introduce GDA approaches in more detail based on this separation as it provides better differentiation of the methodologies and improved readability. \cref{tab:methods} shows a summary of GDA techniques, categorized following this taxonomy and the methods' methodologies.

\section{Rule-based Approaches for GDA}
\label{sec:simpleaug}

\begin{table}[!th]
\scriptsize
\caption{A summary of GDA techniques, categorized by whether they are learned augmentations and their methodologies.}
\label{tab:methods}
\centering
\begin{tabular}{l|l|l|ccc|ccc}
\toprule
\multicolumn{1}{c}{} & \multirow{2}{*}{Methodology} & \multirow{2}{*}{Representative Works} & \multicolumn{3}{|c}{Task Level} & \multicolumn{3}{|c}{Augmented Data}  \\
\multicolumn{2}{c|}{} & & Node & Graph & Edge & Structure & Feature & Label \\
\midrule
\multirow{25}{*}{Rule-based GDA}
& \multirow{7}{*}{Stochastic Dropping/Masking}
& DropEdge~\cite{rong2019dropedge} & \cmark &  &  & \cmark &  &  \\
& & DropNode~\cite{feng2020graph} &  & \cmark &  &  & \cmark &  \\
& & NodeDropping~\cite{you2020graph} &  & \cmark &  & \cmark &  &  \\
& & Feature Masking~\cite{thakoor2022largescale} &  &  &  &  & \cmark &  \\
& & Feature Shuffling~\cite{velickovic2019deep} &  &  &  &  & \cmark &  \\
& & DropMessage~\cite{fang2022dropmessage} &  &  &  &  & \cmark &  \\
& & Subgraph Masking~\cite{you2020graph} &  &  &  & \cmark & \cmark &  \\
\cmidrule(lr){2-9}
& \multirow{3}{*}{Subgraph Cropping/Substituting}
& GraphCrop~\cite{wang2020graphcrop} &  & \cmark &  & \cmark &  &  \\
& & M-Evolve~\cite{zhou2020data} &  & \cmark &  & \cmark &  &  \\
& & MoCL~\cite{sun2021mocl} &  & \cmark &  & \cmark & \cmark &  \\
\cmidrule(lr){2-9}
& \multirow{4}{*}{Mixup}
& Graph Mixup~\cite{wang2021mixup} & \cmark & \cmark &  &  &  & \cmark \\
& & ifMixup~\cite{guo2021intrusion} &  & \cmark &  & \cmark & \cmark & \cmark \\
& & Graph Transparent~\cite{park2021graph} &  & \cmark &  & \cmark & \cmark & \cmark \\
& & G-Mixup~\cite{han2022G} &  & \cmark &  & \cmark & \cmark & \cmark \\
\cmidrule(lr){2-9}
& \multirow{3}{*}{SMOTE}
& GraphSMOTE~\cite{zhao2021graphsmote} & \cmark &  &  &  & \cmark &  \\
& & GATSMOTE~\cite{liu2022gatsmote} & \cmark &  &  & \cmark &  &  \\
& & GNN-CL~\cite{li2022graph} & \cmark &  &  & \cmark & \cmark &  \\
\cmidrule(lr){2-9}
& \multirow{1}{*}{Diffusion}
& GDA~\cite{klicpera2019diffusion} & \cmark &  &  & \cmark &  & \\
\cmidrule(lr){2-9}
& \multirow{1}{*}{Counterfactual Augmentation}
& CFLP~\cite{zhao2021counterfactual} &  &  & \cmark & \cmark &  & \cmark \\
\cmidrule(lr){2-9}
& \multirow{2}{*}{Attribute Augmentation}
& LA-GNN~\cite{liu2021local} & \cmark &  &  &  & \cmark &  \\
& & SR+DR~\cite{song2021topological} & \cmark &  &  &  & \cmark &  \\
\cmidrule(lr){2-9}
& \multirow{2}{*}{Pseudo-labeling}
& Label Propagation~\cite{zhu2005semi} & \cmark &  &  &  &  & \cmark \\
& & PTA~\cite{dong2021equivalence} & \cmark &  &  &  &  & \cmark \\
\midrule
\multirow{16}{*}{Learned GDA}
& \multirow{5}{*}{Graph Structure Learning}
& GAug~\cite{zhao2021data} & \cmark &  &  & \cmark &  &  \\
& & GLCN~\cite{jiang2019semi} & \cmark  &  &  & \cmark &  &  \\
& & LDS~\cite{franceschi2019learning} & \cmark  &  &  & \cmark &  &  \\
& & ProGNN~\cite{jin2020graph} & \cmark &  &  & \cmark &  &  \\
& & Eland~\cite{zhao2021action} & \cmark &  &  &  \cmark &  &  \\
\cmidrule(lr){2-9}
& \multirow{4}{*}{Graph Adversarial Training}
& RobustTraining~\cite{xu2019topology} & \cmark &  &  & \cmark &  &  \\
& & AdvT~\cite{dai2019adversarial} &\cmark  &  &\cmark  &\cmark  &  &  \\
& & FLAG~\cite{kong2022robust} & \cmark & \cmark &\cmark  &  & \cmark  &  \\
& & GraphVAT~\cite{feng2019graph} &\cmark  &  &  & & \cmark  &  \\
\cmidrule(lr){2-9}
& \multirow{2}{*}{Graph Rationalization}
& DIR~\cite{wu2021discovering} &  & \cmark &  & \cmark & \cmark &  \\
& & GREA~\cite{liu2022graph} &  & \cmark &  & \cmark & \cmark &  \\
\cmidrule(lr){2-9}
& \multirow{3}{*}{Automated Augmentation}
& AutoGDA~\cite{zhao2022autogda} & \cmark &  &  & \cmark & \cmark &  \\
& & GraphAug~\cite{luo2022automated} &  & \cmark &  & \cmark & \cmark &  \\
& & JOAO~\cite{you2021graph} &  & \cmark &  & \cmark & \cmark &  \\
\bottomrule
\end{tabular}
\end{table}

Owing to their simplicity and efficiency, rule-based graph data augmentation methods are the most commonly used augmentation techniques in graph machine learning. The rule-based GDA approaches can generally categorized into three categories, where the first category of methods would remove part of the data (e.g., stochastic masking) to create new graph data, the second category of methods creates new graph data by creating additional data (e.g., Mixup, Pseudo-labeling), and the third category includes methods that manipulate the data following rules can involve both removing and adding operations (e.g., Diffusion, etc.) In the following subsections, we summarize the representative approaches in each category.

\subsection{Data Removal}
\label{sec:dataremove}

\paragraph{Edge Dropping} Edge dropping methods stochastically remove a certain fraction of edges from the graph data. Aiming to alleviate the known over-smoothing problem of GNNs, \citet{rong2019dropedge} first proposed DropEdge which randomly dropped a fixed fraction of edges in each training epoch, resembling Dropout \cite{srivastava2014dropout}. More specifically, at the beginning of each training epoch, the modified adjacency matrix $\tilde{\mathbf{A}}$ is defined by
\begin{equation}
    \tilde{\mathbf{A}} = \mathbf{M} \odot \mathbf{A}, 
\end{equation}
where $\mathbf{M} \in \{0, 1\}^{N \times N}$ is a binary mask on the adjacency matrix s.t. $M_{i,j} = Bernoulli(\varepsilon)$, $\varepsilon \in (0,1)$ is the drop rate hyper-parameter, and $\odot$ denotes the Hadamard product.

During GNN training, DropEdge adopts a newly sampled $\tilde{\mathbf{A}}$ instead of the original graph structure $\mathbf{A}$ for message passing (e.g., \cref{eq:gcn}) in each training epoch. By showing the GNN models different perturbations of the graph in each training epoch, DropEdge improves the model's generalization and shows significant performance improvements on deeper GNNs, indicating that the strategy mitigates over-smoothing. Several other methods~\cite{you2020graph,thakoor2022largescale,zhao2022autogda} also adopt random edge masking in other learning schemes such as self-supervised learning, which conducts the same operation as DropEdge.

\paragraph{Node Dropping} Similar to edge dropping, node dropping methods stochastically remove nodes from the graph. Node dropping is typically implemented in two ways: removing all features of the target nodes from the feature matrix, or removing the target nodes along with all the edges connected with them from the graph structure. 
\citet{feng2020graph} proposed DropNode, which follows the first schema. Concurrently, \citet{you2020graph} proposed NodeDropping following the latter.

Both DropNode~\cite{feng2020graph} and NodeDropping~\cite{you2020graph} aim to randomly remove a fraction of the nodes from the given graph, assuming that the missing nodes should not affect the semantic meanings of the remaining nodes, or the whole graph $G$. 
\citet{feng2020graph} focused on semi-supervised node classification, where a consistency loss is used on the predicted logits of different augmented versions of the graphs. On the other hand, \citet{you2020graph} focused on self-supervised graph representation learning with contrastive targets.

\paragraph{Feature Masking} Other than the graph structure, i.e., nodes and edges, multiple works also adopted masking augmentations on the node features. For example, graph contrastive learning methods~\cite{thakoor2022largescale,you2020graph,you2021graph,zhu2020deep} commonly utilize stochastic feature masking as an efficient way of augmenting or corrupting the graph. On top of randomly masking feature values (i.e., random entries in $\mathbf{X}$) or feature signals (i.e., random columns in $\mathbf{X}$), \citet{velickovic2019deep} utilized row swapping as an effective way of corrupting the graph. Specifically, \citet{velickovic2019deep} randomly re-assigned the each node's feature vector to another node in the graph, which can be obtained by row-wise shuffling of $\mathbf{X}$.

More recently, \citet{fang2022dropmessage} proposed DropMessage, which masks the features aggregated by message passing in GNNs. More specifically, denoting the aggregated neighbor feature of node $v$ by the $l$-th layer as $\boldsymbol{h}_{\mathcal{N}(v)}^l$ (\cref{eq:gnn}), DropMessage randomly applies a binary mask on $\boldsymbol{h}_{\mathcal{N}(v)}^l$ for each node $v \in \mathcal{V}$ in every GNN layer. Similar to other dropping methods, the masks are sampled according to a Bernoulli distribution.

\paragraph{Subgraph Cropping} Another common data removal augmentation approach is cropping out part of the graph data. Such subgraph cropping can usually be achieved by either sampling the remaining subgraph or the subgraph that will be cropped out. For example, \citet{you2020graph} first proposed the Subgraph augmentation, which samples the remaining subgraph via random walk. The method later learns the graph representations by contrasting the sampled subgraphs, with the assumption that the semantics of the whole graph can be preserved in part or its local structure. On the other hand, GraphCrop~\cite{wang2020graphcrop} crops a contiguous subgraph from each of the given graph object. GraphCrop adopts a graph diffusion-based node-centric strategy, performing graph diffusion on the randomly selected seed nodes, to maintain the topology characteristics of original graphs after the cropping.


Other than supervised graph representation learning, the above-discussed stochastic data removal methods are also commonly used in self-supervised graph representation learning methods as an efficient way of augmenting/corrupting graph data. For example, several methods~\cite{velickovic2019deep,you2020graph,you2021graph,thakoor2022largescale} use one or multiple of the above-mentioned techniques as augmentation methods for generating the augmented views of graph data. We further elaborate on the usage of data removing augmentations for self-supervised learning in \cref{sec:application}.

\subsection{Data Addition}
\label{sec:dataadd}

\paragraph{Data Interpolation}
With it's simplicity and effectiveness, Mixup~\cite{zhang2017mixup} has been commonly used in image and language domains for augmenting new data samples. Specifically, Mixup constructs virtual training examples by interpolating two labeled training samples:
\begin{equation}
    \begin{array}{cc}
         & \tilde{\boldsymbol{x}} = \lambda\boldsymbol{x}_i + (1-\lambda)\boldsymbol{x}_j, \\
         & \tilde{\boldsymbol{y}} = \lambda\boldsymbol{y}_i + (1-\lambda)\boldsymbol{y}_j,
    \end{array}
\end{equation}
where $(\boldsymbol{x}_i, \boldsymbol{y}_i)$ and $(\boldsymbol{x}_j, \boldsymbol{y}_j)$ are two randomly selected labeled training examples, and $\lambda \in [0,1]$. By linearly interpolating the feature vectors and labels, Mixup incorporates the prior knowledge and extends the training distribution. Similarly, Manifold Mixup~\cite{verma2019manifold} performs Mixup on latent intermediate representations instead of raw features of the two training samples.  

The direct analog of Mixup on graphs is not obvious, given the inter-dependent and irregular nature of graph data.
\citet{verma2019graphmix} proposed GraphMix that augmented the training of a GNNs with a Fully-Connected Network, which is trained by interpolating the hidden states and labels. 
As GraphMix is more of a regularization method than the analog of Mixup on graphs, \citet{wang2021mixup} proposed Graph Mixup, which analogized Manifold Mixup with a two-branch graph convolution module. Given a pair of nodes, Graph Mixup mixes their raw features, passes them into the two-branch GNN layer, and mixes the hidden representations of each layer. Notably, mixing up the nodes on features and hidden states avoids re-assembling the local neighborhoods of the two nodes. Graph Mixup also works for the task of graph classification.
To avoid the node matching problem when mixing up two independent graphs, Graph Mixup mixes the latent representations of the pair of graphs. 

On the other hand, ifMixup~\cite{guo2021intrusion} directly applies Mixup on the graph data instead of the latent space for graph-level tasks. As the pair of graphs are irregular and the nodes from two graphs are not generally aligned, ifMixup arbitrarily assigns indices to the nodes in each graph and matches the nodes according to the indices. 
Empirically, ifMixup shows marginal performance improvements over Graph Mixup on the task of graph classification.
Following ifMixup, Graph Transplant~\cite{park2021graph} also mixes graph in data space. Unlike ifMixup which randomly matches nodes during mixing, Graph Transplant uses substructures as mixing units to preserve the local structural information. Graph Transplant employs the node saliency information to select one meaningful substructure from each graph, where the saliency information is defined as the $l_2$ norm of the gradient of the classification loss. 

Different from the above Mixup-based methods which operate on instance level, \citet{han2022G} proposed G-Mixup that performs Mixup on class-level. Instead of directly interpolating the individual graphs, G-Mixup interpolates the graph generators (graphons) for each class. Specifically, G-Mixup first estimates a graphon for each class of the training graphs, then mixes up the graphons of different classes, and finally generate synthetic graphs with the mixed graphons. Denoting the graphons of classes $a$ and $b$ as $W_a$ and $W_b$, respectively, G-Mixup can be formulated as
\begin{equation}
    \begin{array}{cc}
         \tilde{\boldsymbol{x}} \sim W_c, \text{ where}\quad W_c = \lambda W_a + (1-\lambda)W_b, \\
         \tilde{\boldsymbol{y}} = \lambda\boldsymbol{y}_a + (1-\lambda)\boldsymbol{y}_b,
    \end{array}
\end{equation}
where $\boldsymbol{y}_a$ and $\boldsymbol{y}_b$ are corresponding labels for graphs in classes $a$ and $b$, respectively.

Besides Mixup, SMOTE~\cite{chawla2002smote} is also a classical data augmentation method that interpolates data instances. Different from Mixup which interpolates examples from different classes, SMOTE interpolates examples within the minority classes. Hence, SMOTE is especially effective when dealing with imbalanced data. On graph data, GraphSMOTE~\cite{zhao2021graphsmote} augments the minority class by over-sampling synthetic nodes and then generating edges for them. GATSMOTE~\cite{liu2022gatsmote} and GNN-CL~\cite{li2022graph} further utilize attention designs to improve the edge generating process between the synthetic nodes and original nodes in the graph.

\paragraph{Counterfactual Augmentations}
Counterfactual augmentation has been relatively under-explored in the field of graph machine learning.
\citet{zhao2021counterfactual} first proposed a counterfactual data augmentation method CFLP for the task of link prediction. To better understand the relationship between observed graph structure and link formation, CFLP asks the counterfactual question of ``would the link still exist if the graph structure became different from observation?'' To answer the question, \citet{zhao2021counterfactual} proposed counterfactual links that approximates the unobserved outcome in the question. CFLP then trains a link prediction model with both the given training data and the generated counterfactual links (as augmented data). Similarly, CLBR~\citet{zhu2022data} proposed counterfactual data augmentation for bundle recommendation. CLBR generates the counterfactual example by answering the counterfactual question ``what would a user interact with if the bundle-item affiliation relations change?''.

\paragraph{Attribute Augmentation}
Besides updating the graph topology, several works were also proposed to augment the graph data by generating additional node attributes. For example, LA-GNN~\cite{liu2021local} enhances the locality of node representations by generating node features based on the conditional distribution of the local structures and neighbor features. LA-GNN learns the new features of each node by the conditional distribution of its local neighborhood. The generated feature is directly used together with the raw node features as part of the input of GNNs for both training and inference. Similarly, SR+DR~\cite{song2021topological} generates topology features with DeepWalk~\cite{perozzi2014deepwalk}, and uses a dual GNN model with topology regularization to jointly train with both raw and topology features.

\paragraph{Pseudo-labeling}
The training data in graph tasks is often only partially labeled due to the generally high cost of human labeling. With the large amount of unlabeled data, pseudo-labeling for the unlabeled data is often adopted under semi-supervised learning settings. Label Propagation~\cite{zhu2002learning,zhu2005semi,dong2021equivalence} is one of the most classical methods for generating pseudo labels when only part of the nodes in the graph are labeled. Label propagation assumes that the two nodes are more likely to have the same label if they are connected, so it iteratively propagates node labels along the edges. With the propagated labels on the previously unlabeled nodes, the GNN model can then be trained with more labeled data. 

\subsection{Data Manipulation}
\label{sec:datamani}
\paragraph{Diffusion}
\citet{klicpera2019diffusion} first proposed generalized graph diffusion that modeled a ``future'' state of the graph where the signals were more spread out. Specifically, the generalized graph diffusion is formulated as
\begin{equation}
    \tilde{\mathbf{A}} = \sum_{k=0}^{\infty}\theta_k\mathbf{T}^k,
\end{equation}
where $\theta_k$ denote the global-local coefficient and $\mathbf{T} \in \mathbb{R}^{N\times N}$ represents the transition matrix derived from the adjacency matrix $\mathbf{A}$ (e.g., $\mathbf{A}\mathbf{D}^{-1}$ or $\mathbf{D}^{-\frac{1}{2}}\mathbf{A}\mathbf{D}^{-\frac{1}{2}}$). 
$\theta_k$ is usually pre-defined by specific diffusion variants, e.g., heat kernel~\cite{kondor2002diffusion} ($\theta_k=e^{-t}\frac{t^k}{k!}$) or Personalized PageRank (PPR)~\cite{page1999pagerank} ($\theta_k = \alpha(1-\alpha)^k$), where $\alpha$ denotes the teleport probability in a random walk and $t$ is diffusion time. The analytical solution of the heat kernel and PPR diffusions are defined as
\begin{equation}
    \tilde{\mathbf{A}}^{\text{heat}} = e^{-(t\mathbf{T} - t)}; \quad
    \tilde{\mathbf{A}}^{\text{PPR}} = \alpha (\mathbf{I}_N - (1-\alpha)\mathbf{T})^{-1},
\end{equation}
where $\mathbf{I}_N$ is the $N$ by $N$ identity matrix. As the obtained adjacency matrix after diffusion $\tilde{\mathbf{A}}$ is often too dense as input for GNNs, graph sparsification is commonly conducted to filter out some trivial edges, e.g., setting a threshold to cut-off edges with small weights.

For (semi-)supervised learning on graphs, $\tilde{\mathbf{A}}$ can be directly used for both training and inferencing with GNNs \cite{klicpera2019diffusion}. While most message passing-based GNNs are only capable of aggregating one-hop information in each layer, the augmented graph after diffusion allows GNNs to learn from multi-hop (global) information without specifically re-designing the GNN models. In self-supervised graph representation learning, $\tilde{\mathbf{A}}$ is often used as the augmented view for self-supervised learning objectives such as contrastive learning~\cite{hassani2020contrastive,yuan2021semi}.

\paragraph{Subgraph Substituting}
Several methods also make use of special substructures such as motifs and functional groups during subgraph augmentation. For example,
M-Evolve~\cite{zhou2020data} utilizes motifs to augment the graph data. M-Evolve first finds and selects the target motif in the graph, then adds or removes edges within the selected motifs based on a sampling weight calculated with Resource Allocation index.
Similarly, MoCL~\cite{sun2021mocl} utilizes biomedical domain knowledge to augment the molecular graphs on the substructures such as functional groups. 
MoCL selects a substructure from each molecular graph and replaces it with another substructure. 

\section{Learned Approaches for GDA}
\label{sec:learnedaug}

In the previous section, we introduced rule-based GDA approaches where no learnable parameters are involved during data augmentation. However, these approaches could sometimes be suboptimal since the augmentations do not take advantage of the rich information from downstream tasks, especially in (semi-)supervised training. Indeed, some prior works from the vision \cite{cubuk2019autoaugment} and natural language \cite{niu2019automatically} learning domains show the promise of learned augmentation approaches. To address this concern, learned GDA approaches are proposed to learn augmentation strategies in a data-driven manner. The existing methods can be categorized into the following types: (1) structure learning, (2) adversarial training, (3) rationalization, and (4) automated augmentation.

\subsection{Graph Structure Learning}
\label{sec:gsl}
In real-world scenarios, given graph structures are often incomplete~\cite{franceschi2019learning}, noisy~\cite{jin2020graph,luo2021learning} or manipulated by adversarial attacks~\cite{jin2020adversarial,gunnemann2022graph}. Simply applying rule-based GDA approaches for training (semi-)supervised models on such graphs can lead to suboptimal performances, as they may not necessarily generate better graph structures for downstream tasks. To tackle these issues, several works propose graph structure learning approaches which aim to search for a better graph structure that augments the initial graph structure.
Essentially, those methods treat the graph structure as learnable parameters and iteratively refine it while learning the model parameters~\cite{zhao2021data,jin2020graph,franceschi2019learning,chen2020iterative,luo2021learning,zheng2020robust}. Numerous studies have demonstrated the effectiveness of graph structure learning methods in improving model generalization~\cite{zhao2021data,chen2020iterative} and robustness~\cite{jin2020graph,zheng2020robust}. In the following, we introduce several representative works that fall into the category of graph structure learning. 


\paragraph{Improving Generalization} There are numerous methods for graph structure learning that target improving the generalization performance. Overall, they can be divided into two categories based on the adjacency matrix which they learn: learning \emph{continuous structure} and learning \emph{discrete structure}.

Although the original adjacency matrix is usually discrete (or binary),  \emph{continuous structure} methods do not assume the learned adjacency matrix to be discrete, because modeling discrete structure requires additional efforts in optimization. 
Typically, these methods either model the adjacency matrix as free parameters or use a parameterized neural network to model the structure. For instance, GLCN~\cite{jiang2019semi} is an early work which proposes a unified network architecture to learn an optimal graph structure and GNN. It incorporates the similarities of node features to learn a sparse and continuous graph structure. Formally, it defines a graph learning loss $\mathcal{L}_\text{GL}$ as follows:
\begin{equation}
   \mathcal{L}_\text{GL} = \sum_{i, j=1}^N\left\|{\bf x}_i-{\bf x}_j\right\|_2^2 \tilde{\bf A}_{i j}+\gamma\|\tilde{\bf A}\|_F^2 + \beta \|\tilde{\bf A} - {\bf A}\|_F,
\end{equation}
where the first two terms control the smoothness and sparsity of the augmented graph, respectively; the third term forces the augmented graph to be close to the original graph; $\alpha$ and $\beta$ are the hyper-parameters that balance the three terms. By minimizing the graph learning loss together with the classification loss, GLCN is able to learn a graph structure that best serves the downstream task.
Similar to GLCN, TO-GCN~\cite{yang2019topology} also considers the feature similarity, but it further employs label similarity to refine the graph topology.  To handle the inductive learning setting, IDGL~\cite{chen2020iterative} casts the graph structure learning problem as similarity metric learning  which will be jointly trained with the prediction model dedicated to a downstream task. 
To encourage learning graph structure invariant to task-irrelevant information, \citet{sun2022graph} utilized the Information Bottleneck~\cite{tishby2000information} principle  to solve the graph structure learning problem. Moreover, SLAPS~\cite{fatemi2021slaps} identifies a supervision starvation problem in previous structure learning approaches and proposes to incorporate additional self-supervision by designing a feature denoising task.

Despite the appeal of the first type of methods, continuous structures typically deviate from the original, sparse and \emph{discrete structure} evident in many real-world graphs.   
 To address this concern, some works focus on sampling the graph structures from a targeted distribution. 
For instance, by taking advantage of neural edge predictors like GAE~\cite{kipf2016variational}, \citet{zhao2021data} proposed GAug to generate plausible edge augmentations for an input graph. The output of the edge predictor can be formulated as 
\begin{equation}
\mathbf{M}=\sigma_0\left(\mathbf{Z Z}^T\right), \text { with } \mathbf{Z}= {\bf H}^{l},
\end{equation}
where ${\bf M}$ is the edge probabilities matrix and $\sigma_0$ is an element-wise sigmoid function.
Based on the edge probabilities matrix, two variants GAug-M and GAug-O are proposed to tackle augmentation in settings where edge manipulation is and is not feasible at inference time, respectively. Specifically, GAug-M deterministically adds edges with the highest edge probabilities to the graph at inference time; GAug-O optimizes the graph structure by minimizing the downstream classification loss together with the edge prediction loss and samples the adjacency matrix according to an element-wise Bernoulli distribution.
Another representative work is LDS~\cite{franceschi2019learning}, which aims at learning discrete structure between data points while learning GNN parameters. It models the process as learning the edge probability matrix, which parameterizes the element-wise Bernoulli distribution from which the discrete structure is sampled. Then it formulates the learning process as a bi-level problem and updates the structure and model parameters in a differentiable way. 
Further work ~\cite{shang2021discrete} improves the efficiency of LDS by converting the bi-level problem to a uni-level problem and extends it to multivariate time series.
In addition to Bernoulli distribution, recent studies have investigated other distributions to sample the discrete structure. For example, to account for the underlying generation of graphs, GEN~\cite{wang2021graph} hypothesizes that the estimated graph is drawn from Stochastic Block Model (SBM)~\cite{holland1983stochastic} and learns it in an end-to-end manner. Similarly, BGCN~\cite{zhang2019bayesian} iteratively trains an assortative mixed membership stochastic block model with predictions of GCN to produce multiple denoised graphs, and ensembles results from multiple GCNs. To explicitly guarantee the strength and diversity of graph augmentation, MH-Aug~\cite{park2021metropolis} draws augmented graphs from an explicit target distribution through the Metropolis-Hastings algorithm, which can also be viewed as a  graph structure learning process.

Instead of drawing discrete structures from targeted distributions, another line of works focus on dropping/adding edges from the original graph which can also lead to a discrete adjacency matrix. For instance, to improve the performance of GNNs under random noise, PTDNet~\cite{luo2021learning} proposes to prune task-irrelevant edges by penalizing the number of edges in the sparsified graph and imposing the low-rank constraint with parameterized networks.  Similarly, NeuralSparse~\cite{zheng2020robust} also learns to drop task-irrelevant edges; it takes node/edge features as parts of input and jointly optimizes graph sparsification from the supervision of downstream task. Moreover, \citet{gao2021training} proposed TADropEdge which leverages the graph spectrum to generate edge weights that represent the edges' criticality for the graph connectivity and drops edges by treating their weights as probabilities. Besides node classification tasks, \citet{spinelli2021fairdrop} proposed FairDrop for the task of fair graph representation learning, which biasedly dropped edges with a sensitive attribute homophily mask to protect against unfairness. In addition, \citet{chen2020measuring} also proposed AdaEdge, which iteratively adds/removes edges according to the node classification prediction. In each iteration, after the GNN model is sufficiently trained, AdaEdge adds edges between nodes that are predicted to be in the same class with high confidence, and vice versa. AdaEdge iteratively performs GNN training and graph modification until convergence. Besides, \citet{zhao2021action} proposed Eland for the task of anomaly detection on time-stamped user-item bipartite graphs. Eland first transforms the user-item graph into users' action sequences and adopts seq2seq model for future action prediction. The predicted user actions are added back into the graph to yield the augmented graph data. As the augmented graph contains richer user behavior information, Eland enhances the anomaly detection performance and detects anomalies at an early stage.
It is worth mentioning that the aforementioned techniques are focused on one specific task such as node classification. To make graph structure learning benefit various downstream tasks, \citet{liu2022towards} proposed an unsupervised approach to learn graph structures with the aid of self-supervised contrastive learning~\cite{zhu2021graph}.  

While existing methods majorly focus on training-time augmentation, i.e., modifying the training graph data, a new line of work (e.g., GTrans~\cite{jin2022empowering}) introduces test-time augmentation by transforming the test graph through optimizing a self-supervised loss. It has been demonstrated to significantly improve the generalization performance of GNNs on out-of-distribution data.

\paragraph{Improving Robustness} Recent studies have demonstrated the vulnerability of GNNs under adversarial attacks, i.e., carefully-crafted small perturbation on the input graph leads GNNs into giving wrong predictions~\cite{zugner2018adversarial,dai2018adversarial,zugner2019adversarial,jin2021adversarial}. A series of works are proposed to focus on enhancing the robustness of graph neural networks under adversarial attacks by learning clean graph structure. \citet{jin2020graph} observed that adversarial attacks violate important graph properties such as sparsity, low-rank, and feature smoothness; it then proposes the ProGNN framework to robustify GNNs by alternatively updating the graph structure by preserving these graph properties by adding penalizing regularization terms and training GNN parameters on the updated graph structure. Specifically, it defines the following graph learning loss:
\begin{equation}
\mathcal{L}_\text{GL}=\alpha\|\tilde{\bf A}\|_1+\beta\|\tilde{\bf A}\|_* +\lambda \operatorname{tr}\left(\mathbf{X}^T \tilde{\mathbf{L}} \mathbf{X}\right) + \|\tilde{\bf A} - {\bf A}\|_F^2,
\label{eq:prognn}
\end{equation}
where $\|\cdot\|_1$ is the $\ell_1$ norm, $\|\cdot\|_*$ is the nuclear norm, and $\tilde{\bf L}$ is the normalized Laplacian matrix of $\tilde{\bf A}$. The first three terms in \cref{eq:prognn} force the learned graph to preserve the properties of sparsity, low-rank, and feature smoothness, respectively. Similar to GLCN~\cite{jiang2019semi}, ProGNN also includes the downstream classification loss in the graph learning process. Despite the  robustness of ProGNN, it is computationally expensive with O($N^3$) time complexity and O$(N^2)$ space complexity. To speed up ProGNN, LRGNN~\cite{xu2021speedup} decouples the adjacency matrix into a low-rank component and a sparse one, and learns the graph structure by minimizing the rank of the low-rank component and suppressing the sparse one. Furthermore, as robust GNNs tend to yield unsatisfying performance when trained with limited labeled nodes, \citet{dai2022towards} took advantage of self-supervision and uses node attributes to predict the links so as to boost robust performance, which also saves computational cost from direct structure learning. Also using a link predictor, DefenseVAE~\cite{zhang2020defensevgae} employs variational graph autoencoder~\cite{kipf2016variational} to reconstruct graph structure that can reduce the effects of adversarial perturbations and boost the performance of GNNs under adversarial attacks. In addition, utilizing information theory, CoGSL~\cite{liu2022compact} targets at learning the most compact structure relevant to downstream tasks in order to achieve a better balance between robustness and accuracy. Instead of explicitly learning the graph structure, GNNGuard~\cite{zhang2020gnnguard} mitigates the negative effects of  adversarial attacks by assigning higher weights to edges connecting similar nodes while pruning edges between dissimilar nodes, which can also be considered as implicit graph structure learning. While the aforementioned techniques such as ProGNN and GNNGuard have shown robustness in some specific settings, one recent work~\cite{mujkanovic2022defenses} revealed that their robustness decreases significantly under proper evaluation (in particular the adaptive attacks). This suggests that a more powerful and adaptive GSL method is needed for effective defense. 


It is worth noting that there are some other graph structure learning 
 works which aim at learning graphs to improve the scalability of graph machine learning models~\cite{jin2022graph, jin2022condensing,liu2022graphc}. They do not target improving model performance or robustness of GNNs, and hence are not in the GDA scope tackled in this work. 

\subsection{Graph Adversarial Training}
\label{sec:adv}
Adversarial training is a widely used countermeasure for adversarial attacks on computer vision~\cite{goodfellow2014explaining}, and has also been extended to graph domain~\cite{dai2018adversarial,feng2019graph,deng2019batch,hu2021robust,dai2019adversarial,chen2020smoothing,kong2022robust}.  Unlike graph structure learning, graph adversarial training does not seek to find an optimal graph structure. Instead, it augments input graphs with adversarial patterns during model training by perturbing node features or graph structure. The adversarially trained models are expected to tolerate adversarial perturbations in graph data and yield better generalization and robustness performance at test time. At the core of adversarial training is the injection of adversarial examples into the training set, with which the trained model can predict the test adversarial examples properly.  Thus, we can adopt this strategy to enhance the robustness of GNNs as follows, 
\begin{equation}
\min _{\Theta} \max _{\Delta_{\bf A} \in \mathcal{P}_{\bf A} \atop \Delta_{\bf X} \in \mathcal{P}_{\bf X}} \mathcal{L}_{\text{train}}\left(g_{\Theta}({\bf A}+\Delta_{\bf A}, {\bf X}+\Delta_{\bf X}) \right),
\label{eq:adv-training}
\end{equation}
where $\mathcal{L}_\text{train}$ denotes the training loss for the downstream task; $\Delta_{\bf A}$ and $\Delta_{\bf X}$ stand for the perturbation on ${\bf A}, {\bf X}$, respectively; $\mathcal{P}_{\bf A}$ and $\mathcal{P}_{\bf X}$ denote the perturbation space. From the bi-level optimization problem in \cref{eq:adv-training}, we can observe that adversarial training generates perturbations that maximize the prediction loss and updates model parameters to minimize the prediction loss. The process of generating perturbations (i.e., ${\bf A}+\Delta_{\bf A}, {\bf X}+\Delta_{\bf X}$) can be viewed as adversarial data augmentation and we can leverage such augmentations to improve the model robustness and generalization.  

To augment the adjacency matrix, \citet{dai2018adversarial} proposed to randomly drop edges during adversarial training without any optimization on the graph data. While this strategy does not bring significant improvement, such cheap adversarial training still shows some improvement in robust classification accuracy. This finding is also in line with that from \citet{zugner2020certifiable}. Instead of randomly dropping edges, \citet{xu2019topology} leveraged projected gradient descent (PGD) to optimize the bi-level problem and generate perturbations on the discrete structure, which achieves significant improvement in robust performance. Similarly, \citet{chen2019can} and~\citet{dai2019adversarial} also used existing adversarial attacks to modify the input graph structure during adversarial training, designed for network embedding methods. Furthermore, \citet{suresh2021adversarial} proposed to generate adversarial graph augmentation by learning to drop edges such that the augmentation can capture the minimal information that is sufficient to classify each graph. 

On the other hand, there are some works focusing on perturbing the input features to serve as adversarial examples. For instance, \citet{feng2019graph} proposed an adversarial training strategy with dynamic regularization, which aims to reconstruct graph smoothness and constrains the divergence between the prediction of the target node and its connected nodes. \citet{deng2019batch} proposed batch virtual adversarial training to promote the smoothness of GNNs and thus defend against adversarial perturbations. Moreover, \citet{kong2022robust} proposed FLAG which utilizes adversarial training to iteratively augment the node features with gradient-based adversarial perturbations and improves the performances of GNNs on node classification, link prediction, and graph classification tasks. In addition, \citet{zugner2019certifiable} studied certifiable robustness of GNNs w.r.t. perturbations of node attributes and propose a robust training scheme inspired by the certificates.
Several other variants of adversarial training on perturbing node features are introduced in~\cite{wang2019graphdefense-adv-training,hu2021robust}.  

\subsection{Rationalization}
\label{sec:rat}
A \emph{rationale} is defined as a subset of input features that best represent, guide and support model prediction~\cite{liu2022graph}. In the graph domain, rationales are usually intrinsically learned subgraphs, as a form of augmented graph data, that are representative, provide information or explanation to the graph models. Rationale subgraphs are either used separately or in addition to the original graph to inform model decisions.
These methods are intrinsically interpretable models that includes a rationalization component in the model, as opposed to post-hoc explanation methods.
Rationalization is commonly applied to graph property prediction or graph classification tasks~\cite{wu2020generalization,liu2022graph, you2020graph, chenlearning, miao2022interpretable, li2022learning} on drug discovery, molecular and polymer datasets, etc. 

Rationalization in the graph domain first appeared as a means to improve interpretability and general graph classification performance.  \citet{yu2020graph} found similarity to the Information Bottleneck (IB) problem, and proposed the Graph Information bottleneck framework (GIB), which learns to generate the maximally informative and compressed subgraph (IB-graph) by leveraging a bi-level optimization scheme and a novel connectivity loss. Also rooted in the IB paradigm, \citet{miao2022interpretable} proposed GSAT to better learn and select task-relevant subgraphs that improve interpretation and prediction by injecting stochasticity into the attention weights in order to constrain information from task-irrelevant components.
GREA, another rationalization work proposed by \citet{liu2022graph}, introduces a new augmentation method called environment replacement. They aim to improve rationale identification by separating rationale and environment (remaining subgraph after rationale identification), and subsequently replacing the environment to generate augmented virtual data examples. 

Rationalization models are also effective at solving data bias and out-of-distribution (OOD) problems for graph property prediction tasks. They not only provide better interpretation but also better generalization. \citet{wu2021discovering} proposed DIR to generate distribution perturbation on training data with causal intervention. Based on the idea that causal patterns are stable to distribution shift, they created a rationale generator that separates causal and non-causal graphs, applies causal intervention to create perturbed distributions, and then jointly learn both the causal and non-causal representation to minimize invariant risk. Similarly, \citet{chenlearning} also took a causal perspective to solve the OOD problem. They proposed CIGA to model the graph generation process and the interactions between invariant and spurious features with Structural Causal Models (SCM). The resulting subgraphs generated by CIGA maximally preserves the invariant intra-class information.
 \citet{li2022learning} also proposed to separate invariant and variant graphs. In their framework GIL, they proposed a GNN based subgraph generator to identify potentially invariant subgraphs, then infer latent environment labels for the variant subgraphs, before jointly optimizing all modules.

\subsection{Automated Augmentation}
\label{sec:autoaug}

GDA techniques mentioned in \cref{sec:simpleaug} take rule-based approaches to augment graph data, applying the same augmentation method to subgraphs and graphs which embody different attributes and characteristics like degree distribution and homophily. To tackle this issue, Automated GDA techniques~\cite{sun2021automated, luo2022automated,zhao2022autogda,you2021graph, kose2022fair, hassani2022learning, zhu2021graph} were recently explored to automatically learn tailored augmentations for different subgraphs or graphs. For example, \citet{sun2021automated} proposed AutoGRL for the task of node classification. Through the training process, AutoGRL learns the best combination of GDA operations, GNN architecture, and hyperparameters. The searching space of AutoGRL includes four GDA operations implemented by random masking and GAug-M~\cite{zhao2021data}: drop features, drop nodes, add edges, and remove edges.

Since automated GDA objectives are often complex to optimize, some recent works use reinforcement learning approaches as a solution. \citet{zhao2022autogda} framed the AutoGDA as a bi-level optimization problem, aiming to find a different set of augmentation strategies for each community in the graph as they observed various characteristics in each community. They employ an RL-agent to generalize the learning and find localize augmentation strategies for node classification tasks. On graph classification tasks, \citet{luo2022automated} set out to learn an automated augmentation model with GraphAug, to provide label-invariant augmentations for each graph in the dataset. Applying reinforcement learning, they maximize the estimated label-invariance probability to learn the augmentation category and transformation selection.

Another group of works on automated augmentation focus on graph contrastive learning. \citet{you2021graph} proposed to learn augmentations to replace ad hoc and handpicked augmentations for contrastive learning. They design an augmentation-aware projection head to avoid complicated augmentations, and formulate a bi-level optimization problem to learn both the augmentation strategy and graph representation. \citet{hassani2022learning} learned a probabilistic policy that contains a set of distributions over different augmentation operations in their method LG2AR, and samples an augmentation strategy from the policy in each training epoch. 
\citet{zhu2021graph} proposed GCA, which proposes adaptive augmentations based on node centrality measures. Unlike the aforementioned methods which find the best augmentation strategy for the dataset, GCA adaptively augments different nodes according to their importance. \citet{wang2021molecular} proposed to use a generative probabilistic model and a learnable feature selector to automatically parameterize topological and attribute augmentations, which can also provide explanations for underlying patterns in molecular graphs. Lastly, \citet{kose2022fair} proposed FairAug which utilizes adaptive augmentations for fair graph representation learning.

\section{GDA for Self-supervised Learning}
\label{sec:application}

\begin{table}[t]
\scriptsize
\caption{Representative self-supervised graph learning works that utilized graph data augmentation techniques.  \\ $^\dag$Although the methods in this category are semi-supervised methods, they used GDA operations with only self-supervised learning objectives (i.e, consistency loss). Therefore, we categorize their GDA techniques as designed for self-supervised learning objectives.}
\label{tab:sslmethods}
\centering
\Scale[1]{\begin{tabular}{l|l|ccc|ccc}
\toprule
\multicolumn{1}{c|}{} & \multirow{2}{*}{Representative Works} & \multicolumn{3}{c|}{Task Level} & \multicolumn{3}{c}{Augmented Data} \\
\multicolumn{1}{c|}{} & \multicolumn{1}{c|}{} & Node & Graph & Edge & Structure & Feature & Label \\
\midrule
\multirow{5}{*}{Contrastive Learning}
& DGI~\cite{velickovic2019deep} & \cmark &  &  &  &  \cmark \\
& GRACE~\cite{zhu2020deep} & \cmark &  &  & \cmark & \cmark &  \\
& MVGRL~\cite{hassani2020contrastive} & \cmark &  &  & \cmark &  \\
& GraphCL~\cite{you2020graph} &  & \cmark &  & \cmark & \cmark \\
& JOAO~\cite{you2021graph} &  & \cmark &  & \cmark & \cmark &  \\
\midrule
\multirow{4}{*}{Non-contrastive Learning}
& CCA-SSG~\cite{zhang2021canonical} & \cmark &  &  & \cmark & \cmark \\
& GBT~\cite{bielak2022graph} & \cmark &  &  & \cmark & \cmark \\
& BGRL~\cite{thakoor2022largescale} & \cmark &  &  & \cmark & \cmark \\
& T-BGRL~\cite{shiao2022link} & \cmark &  &  & \cmark & \cmark \\
\midrule
\multirow{4}{*}{Consistency Training$^\dag$}
& GRAND~\cite{feng2020graph} & \cmark &  &  & \cmark & \cmark \\
& NodeAug~\cite{wang2020nodeaug} & \cmark &  &  & \cmark & \cmark \\
& MV-GCN~\cite{yuan2021semi} & \cmark &  &  & \cmark \\
& NASA~\cite{bo2022regularizing} & \cmark &  &  & \cmark & \\
\bottomrule
\end{tabular}}
\end{table}

Other than directly using the augmented graph data in supervised learning, the most common use case for GDA is under self-supervised learning (SSL) schemes, e.g. contrastive learning. Self-supervised objectives learn representations that are robust to noise and perturbations by maximizing the (dis)agreements of learned representations. Therefore, unlike most of the previously mentioned learned GDA techniques (\cref{sec:learnedaug}) which aim to enhance the task-relevant information in the data, most of the GDA techniques for self-supervised learning are rule-based augmentations (\cref{sec:simpleaug}) which aim to corrupt or perturb the given graph data. Moreover, most self-supervised graph representation learning methods tend to use a combination of several simple GDA operations. In this section, we introduce three commonly used self-supervised graph learning schemes as well as the GDA approaches they utilize.

\subsection{Contrastive Learning}
\label{sec:contrastive}
In recent years, with the rapid development of contrastive learning in CV~\cite{chen2020simple}, many  contrastive learning methods~\cite{zhu2020deep,you2020graph,trivedi2021augmentations,xie2022self,liu2022graph} have been proposed for applications on graph data.
Typically, a graph contrastive learning framework includes three main components: a GDA module that generates different views of the given graph data, a GNN-based encoder to compute the representations, and a contrastive learning objective to train the model. For each data example (nodes for node-level tasks and graphs for graph-level tasks), these methods consider augmented views or variants of itself as associated positive samples and other data examples in the same batch as associated negative samples. Contrastive learning objectives then maximize the (dis)agreements of the representations between each data example with their (negative) positive examples. 

To efficiently generate different augmented data for graph contrastive learning, rule-based data removal operations (\cref{sec:dataremove}) are the most commonly used GDA techniques, as they are fast and easy to apply. For example, multiple methods (GRACE~\cite{zhu2020deep}, GraphCL~\cite{you2020graph}, etc.) adopt stochastic edge dropping and/or feature masking due to their simplicity. DGI~\cite{velickovic2019deep} adopts feature corruption by conducting a row-wise shuffling on the raw node feature matrix $\mathbf{X}$. In general, graph contrastive learning methods usually adopt a combination of multiple augmentation techniques to generate different augmented views. 
GraphCL~\cite{you2020graph} and InfoGCL~\cite{xu2021infogcl} adopt four GDA operations: node dropping which randomly removes nodes along with its edges, edge perturbation which randomly adds or drops edges, attribute masking which randomly masks off certain node attributes, and subgraph sampling which samples connected subgraphs. SUBG-CON~\cite{jiao2020sub} utilizes a subgraph sampler to extract the context subgraph as a proxy of data augmentation.
GRACE~\cite{zhu2020deep} uses only the basic random edge dropping and attribute masking for creating different views of the graph. 

Other than data removal augmentations, graph diffusion is also commonly used in contrastive learning as it can naturally create a ``future view'' of the given graph where the information are more spread out. MVGRL~\cite{hassani2020contrastive} adopts the diffusion graph proposed by GDC~\cite{klicpera2019diffusion} as the second view. Interestingly, \citet{hassani2020contrastive} showed that using three views (original graph, PPR diffusion graph and heat kernel diffusion graph) would not result with better performance than using two views (original graph and one diffusion graph), and concluded ``increasing the number of views does not improve the performance.'' However, \citet{yuan2021semi} later proposed MV-CGC which adopted a similar contrastive learning framework with three views: the original graph, diffusion graph, and a proposed feature similarity view. Empirically, the node representations learned by MV-CGC outperformed those learned by MVGRL on node classification, suggesting that additional well-designed GDA methods or views may be helpful to graph contrastive learning approaches.

More recently, several studies \cite{suresh2021adversarial, trivedianalyzing} pointed out that stochastic rule-based GDA operations may suffer from failing to induce useful task-relevant invariance on common benchmark datasets. Specifically, \citet{trivedianalyzing} analyzed that the generalization error of graph contrastive learning can be bounded under the assumptions of invariance to relevant augmentations, recoverability, and separability, which refer to \textit{data-centric properties}, by instantiating rule-based GDA as a composition of graph edit operations. Such bound demonstrates conditions with low separability and recoverability during the usage of rule-based GDA, which motivates the necessity of inducing task-relevant invariance. 
Following the theoretical analysis, \citet{zhang2022costa} proposed a covariance-preserving feature augmentation technique, in which the augmented feature has bounded variance. \citet{wang2021multi} proposed to use different levels in hierarchical graphs as augmented views.

\subsection{Non-contrastive Learning}
\label{sec:noncontrastive}
While showing promising performance on various tasks, contrastive learning methods rely heavily on disagreement between data examples and their associated negative examples to avoid model collapse~\cite{grill2020bootstrap}. As sampling high quality negative examples is often costly, and random negative sampling usually requires large batch sizes, several works~\cite{grill2020bootstrap,zbontar2021barlow,balestriero2022contrastive} propose non-contrastive self-supervised learning methods to learn representations in a self-supervised manner without needing negative examples. Instead of comparing across different samples, non-contrastive self-supervised methods compare only between different views of the same sample and use designs such as prediction heads and stop gradient to avoid model collapsing~\cite{grill2020bootstrap}, or measure the cross-correlation matrix between the representations learned form different views~\cite{zbontar2021barlow}.

As the non-contrastive methods are designed for more efficient self-supervised learning than the contrastive methods, the GDA techniques they adopt are all the most basic, stochastic ones (\cref{sec:dataremove}). Specifically, all the non-contrastive self-supervised graph representation learning methods (CCA-SSG~\cite{zhang2021canonical}, GBT~\cite{bielak2022graph}, BGRL~\cite{thakoor2022largescale}, and T-BGRL~\cite{shiao2022link}) utilized only random edge dropping and node feature masking as the augmentation strategies. 
While the first three methods generates two augmented views for comparison, to further improve the performance on link prediction under inductive settings, T-BGRL~\cite{shiao2022link} also used the same augmentation strategies but with higher masking probability as an efficient corruption to create an third ``negative'' view to mitigate collapse, which is later used in a triplet loss.

\subsection{Consistency Training}
\label{sec:consistent}

In real GML applications, semi-supervised learning usually plays an important role as only a small fraction of training data are labeled in most of the cases~\cite{wu2020comprehensive}. Due to such label scarcity, consistency training is commonly used to leverage the unlabeled data to improve the model quality. Similar to contrastive learning, consistency training itself is a self-supervised learning objective that aims to maximizes the agreement of representations learned from different views of the data. However, unlike (non-)contrastive learning that compares between data objects, the consistency loss compares the distributions of a batch of representations via metrics like KL-divergence. Therefore, the consistency loss is rarely used itself, but often used along with supervised losses in the semi-supervised learning settings. The final learning objective is usually a linear combination of the supervised loss (e.g., cross entropy for classification tasks) and the consistency loss.

NodeAug~\cite{wang2020nodeaug} uses three local structure-based augmentation operations: replacing attributes, removing and adding edges. NodeAug minimizes the KL-divergence between the node representations learned from the original graph and augmented graph. GRAND~\cite{feng2020graph} creates multiple different augmented graphs with node dropping and feature masking. The consistency loss then minimizes the distances of the representations learned from the augmented graphs. 
NASA~\cite{bo2022regularizing} proposes Neighbor Replace augmentation to randomly replace the 1-hop neighbors with 2-hop neighbors, and then use a neighbor-constrained consistency regularization during training.
To further utilize the information given by different graph diffusions, MV-GCN~\cite{yuan2021semi} generates two complementary views with PPR and heat kernel and learns from both created views and the original graph. Then, it feeds three views of the graph into three GCNs, and uses a consistency regularization loss to reduce the distribution distance of the representations learned across the views, and derives the final node representations as a  combination of the three.

\section{Challenges and Directions}
\label{sec:future}


Despite substantial progress has been achieved in graph data augmentation research, several open problems remain to solve. In this section, we summarize several promising yet under-explored research directions.

\subsection{Domain Adaptation and Regularization}

Given the rapid development of GDA techniques in recent years, automated GDA methods have been proposed to automatically tune the augmentation strategy for different datasets and tasks. 
Nonetheless, the existing automated GDA methods for graph data (as introduced in \cref{sec:autoaug}) mainly focus on specific datasets and downstream tasks. 
Ideally, automated augmentation solutions should be transferable. That is, domain adaptation is a desired characteristic for automated GDA techniques. When the automated augmentation method trained on one dataset could only be used on that dataset, the method may be equivalent to automating the hyperparameter tuning process and lose the generalizability across datasets~\cite{zhao2022autogda}. Therefore, for an ideal automated GDA method, it should be able to be trained on one dataset and used for many, ideally cross domain or under OOD settings. Automated GDA methods that can be transferable across domains are still missing in the literature.
Moreover, on certain types of graph data such as molecule graphs, most commonly used GDA operations would change the underlying semantics of the graph. For example, dropping a carbon atom from the phenyl ring of aspirin breaks the aromatic system and results in a alkene chain~\cite{lee2021augmentation}, which is an entirely different chemical compound. This motivates a need for domain-based regularization methods for such tasks. So far, only \citet{sun2021mocl} proposed MoCL that considers the semantic information brought by local substructures when augmenting the molecule graphs, leaving domain-based regularization GDA methods rather under-explored.


\subsection{Scalability for Large-Scale Graphs}
GDA techniques add additional complexity on top of the existing GNNs, and many GDA techniques use global information during the augmentation process, which might not be able to easily scale. For example, GAug-M~\cite{zhao2021data} involves selecting the top $K$ out of $O(N^2)$ logits for node pairs when selecting edges to add. Such high complexity operations can cause scalability issues in actual applications where the graph size can be very large, e.g., at billion scale.
While complex GDA techniques bring significant performance improvements, the scalability of these methods are still worthy of attention. For example, in order to enable end-to-end training, GAug-O~\cite{zhao2021data} requires back-propagating on the entire learned adjacency matrix, creating massive memory overheads. To improve the performance of DropEdge~\cite{rong2019dropedge}, TADropEdge~\cite{gao2021training} required the pre-calculation of a score for each edge in the graph prior to the training of GNNs. Therefore, to be applicable in practical applications, efficiency is also a necessity for GDA techniques. As mentioned in the previous subsections, automated solution which combine the fast and simple augmentation operations may be a promising direction. Nonetheless, how to design a scalable and efficient automated GDA framework is still an open line of research.


\subsection{Comprehensive Evaluation Criteria and Standards}
Similar to the DA research in other domains, a general concern for GDA research is that the evaluation only focuses on the prediction performance on specific datasets. Although this is likely the most important metric, other metrics such as additional time and resource consumption, transferability, or scalability are also important for researchers to more comprehensively understand the methods. For example, as aforementioned, while graph structure learning methods such as GAug~\cite{zhao2021data} shows promising performances for node classification, the method's design inherently limits its ability to generalize on large-scale graphs. Furthermore, only few works discuss the additional time and resource requirement needed for applying their proposed GDA methods, especially for the learned augmentations which may require training of additional modules. 
Therefore, a set of comprehensive evaluation criteria and standards is desired for better understanding the benefits and costs of the newly proposed GDA methods.
Ideally, such a benchmark could contain multiple datasets in different scales and domains, enabling researchers to better evaluate transferability and scalability tradeoffs.

\subsection{Theoretical Foundation}

GDA is a powerful technology to improve the performance of data-driven inference on graphs without the need of extra labeling effort or complex models. Empirically, GDA methods are also shown to improve the generalization of GML methods and alleviate the over-smoothing problem encountered by GNNs. Yet, there is little rigorous understanding of how and why GDA achieves those benefits, especially for (semi-)supervised learning. Although several works~\cite{zhao2021data,chen2020measuring} have analyzed the relation between graph homophily and classification performance or the over-smoothing problem, there is limited work showcasing rigorous proofs or theoretical bounds on these relationships.

Recently, several works provided theoretical insights of DA in the CV domain. For example, \citet{wu2020generalization} theoretically analyzed the generalization effect of data augmentation on images. They interpreted the effect of data augmentation from a bias-variance perspective, where data augmentation adds new information to model training
while also serving as a regularizer. 
Due to the irregular characteristics of graph data, these theoretical analysis cannot be directly applied for the GDA context. 
Besides the generalization perspective, 
several recent works have studied the certified robustness of GNNs~\cite{zugner2020certifiable}. Improved robustness bounds would be a desired property of GDA techniques. Recent studies~\cite{topping2021understanding} on the topology bottleneck and over-squashing of GNNs provide theoretical guides for edge-based GDA techniques. Counterfactual augmentation methods on graphs such as CFLP~\cite{zhao2021counterfactual} can also bring insights for analyzing GDA from the perspective of causality. 

\section{Conclusion}
\label{sec:conclusion}

Our work presents a comprehensive and structured survey of data augmentation techniques for graph machine learning (GML). We categorized existing graph data augmentation (GDA) techniques three taxonomies from different perspectives, introduced recent GDA approaches based on their core methodology, and introduced their applications in self-supervised learning. Finally, we outlined current challenges as well as directions for future research explorations in the GDA domain. We hope this survey serves as a guide for GML researchers and practitioners to study and use GDA techniques, and inspires additional interest and work on this topic.

\small{
\bibliographystyle{plainnat}
\bibliography{ref}
}
\end{document}